\documentclass[letterpaper, 10pt, conference]{ieeeconf}  
\IEEEoverridecommandlockouts                              
\usepackage{url}
\usepackage{svg}
\usepackage[backend=biber,
            hyperref=true,
            url=false,
            isbn=false,
            doi=false,
            backref=false,
            style=ieee,
            natbib=true,
            mincitenames=1,
            maxcitenames=1,
            citestyle=numeric-comp,
            sorting=nyt,
            block=none]{biblatex}
            
\usepackage{amsmath,amssymb}
\usepackage{graphicx,color}
\usepackage{adjustbox}
\usepackage{booktabs}
\usepackage{multirow}
\usepackage{verbatim}
\usepackage{multicol}
\usepackage{subfig}
\usepackage{caption}
\usepackage{float}
\usepackage{xspace}
\usepackage{gensymb}
\usepackage{marginnote}
\usepackage{siunitx} 
\usepackage{pythonhighlight} 
\graphicspath{{figures/}}
\newcommand{\ba}{\mathbf{a}}

\newcommand{\bs}{\mathbf{s}}

\newcommand{\state}{\mathbf{s}}
\newcommand{\vmax}{v_{\rm max}}

\newcommand{\eg}{e.g.,\xspace}

\newcommand{\dtune}{\mathcal{D}_\textrm{tune}}
\newcommand{\dsim}{\mathcal{D}_\textrm{sim}}
\newcommand{\dreal}{\mathcal{D}_\textrm{phys}}
\newcommand{\pitune}{\pi_\textrm{R2S2R}}
\newcommand{\pisim}{\pi_\textrm{SD}}
\newcommand{\pireal}{\pi_\textrm{RD}}
\newcommand{\polarcasting}{Cast and Pull\xspace}

\DeclareMathOperator*{\argmin}{arg\,min}

\usepackage{pifont}

\definecolor{britishracinggreen}{rgb}{0.23, 0.53, 0.19}

\definecolor{turquoise}{rgb}{0.25, 0.88, 0.82}

\definecolor{navy}{rgb}{0,0,0.5}

\usepackage[font={small}]{caption}

\def\tablename{Table}

\title{\LARGE \bf
Real2Sim2Real: Self-Supervised Learning of Physical \\ Single-Step Dynamic Actions for Planar Robot Casting
}

\author{Vincent Lim$^{1,*}$, Huang Huang$^{1,*}$, Lawrence Yunliang Chen$^{1}$, Jonathan Wang$^{1}$, \\  Jeffrey Ichnowski$^{1}$, Daniel Seita$^{1}$, Michael Laskey$^2$, Ken Goldberg$^1$
\thanks{*Equal contribution.}
\thanks{$^{1}$The AUTOLab at UC Berkeley (automation.berkeley.edu).}%
\thanks{$^{2}$Toyota Research Institute, CA, USA.}
\thanks{Correspondence to: {\tt\scriptsize \{vincentklim, huangr\}@berkeley.edu}}
}
\begin{document}

\maketitle
\thispagestyle{empty}
\pagestyle{empty}

\begin{abstract}
This paper introduces the task of {\em Planar Robot Casting (PRC)}: where one planar motion of a robot arm holding one end of a cable causes the other end to slide across the plane toward a desired target. PRC allows the cable to reach points beyond the robot workspace and has applications for cable management in homes, warehouses, and factories. To efficiently learn a PRC policy for a given cable, we propose {\em Real2Sim2Real}, a self-supervised framework that automatically collects physical trajectory examples to tune parameters of a dynamics simulator using Differential Evolution, generates many simulated examples, and then learns a policy using a weighted combination of simulated and physical data. We evaluate Real2Sim2Real with three simulators, Isaac Gym-segmented, Isaac Gym-hybrid, and PyBullet, two function approximators, Gaussian Processes and Neural Networks (NNs), and three cables with differing stiffness, torsion, and friction. Results with 240 physical trials suggest that the PRC policies can attain median error distance (as \% of cable length) ranging from 8\% to 14\%, outperforming baselines and policies trained on only real or only simulated examples.
Code, data, and videos are available at \url{https://tinyurl.com/robotcast}.
\end{abstract}

\section{Introduction}\label{sec:intro}

Manipulation of deformable objects using a single parameterized dynamic action can be useful for tasks such as fly fishing, lofting a blanket, and playing shuffleboard. Such tasks take as input a desired final state and output one parameterized open-loop dynamic robot action which produces a trajectory toward the final state.  This is especially challenging for long-horizon trajectories with complex dynamics involving friction. Although there is substantial research on learning ballistic throwing or hitting motions for rigid objects~\cite{lynch1999dynamic,zeng_tossing_2019}, there is less research into learning dynamic motions to manipulate deformable objects such as cables and fabrics. 
Dynamics modeling in these contexts is challenging due to uncertainty in deformability, elasticity, and friction during the object's motion. When computing a single action in this setting (without feedback control), the complexities of state estimation and dynamics modeling are compounded by the long duration for which the system evolves after the robot action. Simulation is often used as an alternative to collecting physical examples, which can be time-consuming and hazardous. However, overcoming the simulation to reality (Sim2Real) gap is a long-standing problem in robotics~\cite{rss_workshop, reality_gap_1995,mahler2017dexnet,domain_randomization}, and is particularly challenging for deformable objects in uncertain dynamic environments.

\begin{figure}[t]
\center
\includegraphics[width=0.49\textwidth]{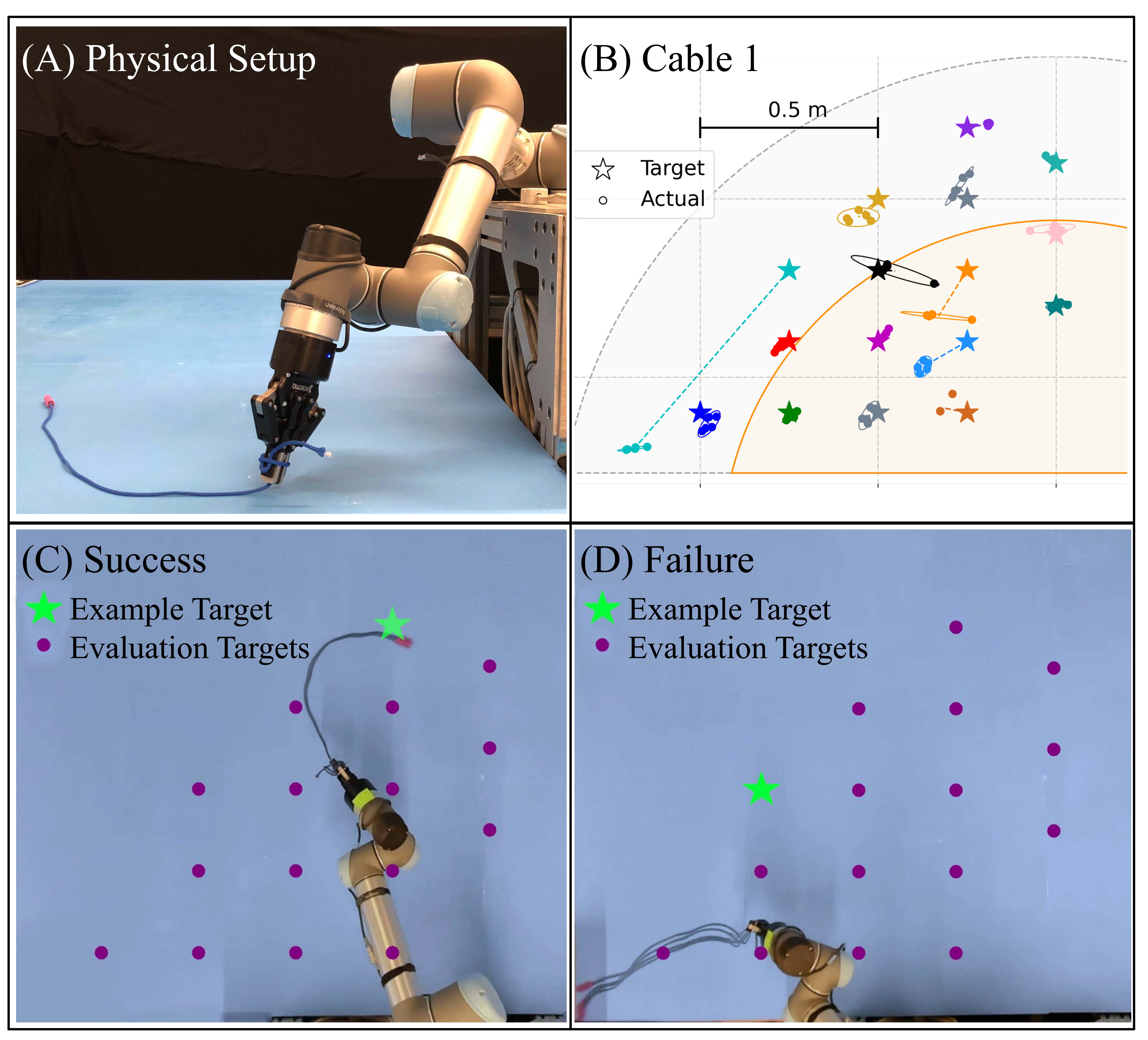}
\caption{
In \emph{Planar Robot Casting} (PRC), a single dynamic planar motion of a robot wrist holding one end of a cable causes the other end of the cable to slide across the plane and stop near a desired target point,  which may lie outside the robot workspace. (A) shows a side view of a UR5 robot with cable and planar workspace, (B) illustrates test performance on Cable 1 for 5 trials (dots) on 16 targets (stars). The gold inner sector represents the robot workspace, while the grey outer sector represents the reachable workspace of the cable. (C,D) each show five superimposed overhead views of the robot and cable with associated target points after PRC actions with the learned policy, in (C) an example with low error and in (D) an example with high error in endpoint position. 
}
\vspace{-16pt}
\label{fig:teaser}
\end{figure}

In this paper, we introduce \emph{Planar Robot Casting} (PRC), where a robot arm dynamically manipulates a cable through 2D space, as illustrated in Fig.~\ref{fig:teaser}. We use the generic term \emph{cable} to refer to any 1D deformable object with low stiffness, such as cables, ropes, and threads. We propose {\em Real2Sim2Real}, a self-supervised robot learning framework that starts by efficiently collecting a small number of physical examples, uses them to tune a simulator, and then uses a combination of physical and simulated examples to train policies for PRC. This paper makes four contributions:

\begin{enumerate}
    \item Real2Sim2Real, a three-step robot learning framework for single-step dynamic actions executed in real environments.
    \item A formulation of the Planar Robot Casting (PRC) problem, with an automated reset motion and parameterized one-step action space that can reach outside of the reachable area of the robot. 
    \item  A physical implementation of PRC using a UR5 robot, and three simulation models of PRC: NVIDIA Isaac Gym \emph{segmented}, Isaac Gym \emph{hybrid}, and PyBullet. Parameter estimation using Differential Evolution~\cite{storn1997differential,2020SciPy-NMeth} with associated code and datasets from 64,350 simulated experiments and 2,076 physical experiments with 3 distinct cables.
    \item An application of Real2Sim2Real to PRC, with results suggesting that Real2Sim2Real can efficiently learn PRC control policies that reach target positions within median error of 14\% of cable length, even when targets are outside the reachable workspace of the robot.
\end{enumerate}

\section{Related Work}\label{sec:rw}

\subsection{Deformable 1D Object Manipulation}

Manipulation of 1D deformable objects such as cables has a long history in robotics; see~\citet{manip_deformable_survey_2018} for a recent survey. Some representative applications include surgical suturing~\cite{robot_heart_surgery_2006,suturing_autolab_2016}, knot-tying~\cite{van_den_berg_2010,case_study_knots_1991,knot_planning_2003,tying_precisely_2016}, untangling ropes~\cite{rope_untangling_2013,grannen2020untangling}, deforming wires~\cite{wire_insertion_1997}, inserting cables into receptacles~\cite{tight_tolerance_insertion_2015,tactile_cable_2020}, and playing diabolo~\cite{diabolo_2020}.  

There is recent interest in using learning-based techniques for manipulating cables, often with pick-and-place actions and quasistatic dynamics, so that the robot deforms the cable while allowing it to settle between actions. This can be combined with learning from demonstrations~\cite{zeng_transporters_2020,seita_bags_2021} and self-supervised learning~\cite{nair_rope_2017,ZSVI_2018}.
Prior work has also used quasistatic simulators~\cite{corl2020softgym} to train cable manipulation policies using reinforcement learning (RL) for transfer to physical robots~\cite{lerrel_2020,yan_fabrics_latent_2020}. 
However, as we show in Sec.~\ref{sec:experiments}, there are limits in the free-end reachability of such quasistatic procedures, motivating high-speed, dynamic motions.

\subsection{Simulator Tuning for Sim2Real}

Sim2Real transfer has emerged as an effective technique wherein a robot can learn a policy in simulation and transfer it to the real world, which can suffer from the mismatch between simulation and real~\cite{reality_gap_1995}. One option for Sim2Real is to perform system identification~\cite{sys_id} to tune a simulator to match real-world data, but this may be challenging with the infinite-dimensional configuration spaces of cables.
An alternative is to use domain randomization over vision~\cite{domain_randomization,cad2rl} or dynamics~\cite{dynamics_randomization_2018} parameters in simulation.
A drawback is that too much randomization can hinder policy training~\cite{sim2real_deform_2018} and the randomization range needs to be hand tuned.
This has motivated work on interleaving policy training using RL in simulation and tuning simulators with real-world data~\cite{closing_sim2real_2019,auto-tune_2021}. Recently, Chang and Padir~\cite{sim2real2sim_2020} proposed a Sim2Real2Sim pipeline for cable unplugging where the Sim2Real step determines grasp points, and the Real2Sim step uses sensors to update a cable model in simulation.
Similarly, we tune a simulator, but do not require repeated iterations of collecting real-world data with simulator tuning, and we train policies using supervised learning and do not need to use potentially brittle RL algorithms~\cite{SpinningUp2018,how_to_train_rl}. Unlike Chang and Padir, we do not require cable sensors and we use the tuned simulator to improve the performance of the physical robot.

\subsection{Dynamic Manipulation}

In dynamic manipulation, a robot executes rapid actions to move objects to desired configurations~\cite{lynch_mason_1993}. 
In early work, Lynch and Mason~\cite{lynch1999dynamic} study planar dynamic manipulation primitives such as snatching, throwing, and rolling, and~\citet{ruiz2011fast} build a physics-based model of pushing actions so that a robot can slide objects into a desired pose.
More recent work has introduced robots that can catch items by predicting item trajectories~\cite{catching_2014}, toss arbitrary objects via self-supervised learning~\cite{zeng_tossing_2019}, 
and swing items upwards using tactile feedback~\cite{swingbot_2020}.
As with these works, we use dynamic motions, but focus specifically on cable manipulation. Furthermore, we use a tuned simulator to accelerate learning and do not require tactile sensing.

\subsection{Dynamic Manipulation of Deformable Objects}

Researchers have developed several analytic physics models to describe the dynamics of moving cables. For example,~\citet{fly_fishing_2004} present a 2D dynamic model of fly fishing. They model the fly line as a long elastica and the fly rod as a flexible Euler-Bernoulli beam, and propose a system of differential equations to predict the movement of the fly line in space and time. 
In contrast to continuum models,~\citet{wang2011analysis} propose using a finite-element model to represent the fly line by a series of rigid cylinders that are connected by massless hinges. In contrast to these works, we evaluate on physical robots.

For robotic dynamic-cable manipulation,~\citet{dynamic_knotting_2010,yamakawa2012simple,high_speed_knotting_2013,one_hand_knotting_tactile_2007} 
simplify the modeling of cable deformation by assuming each cable component follows the robot end-effector motion with constant time delay. 
\citet{kim2016using} study a mobile robot system with a cable attached as a tail that can strike objects.
They use a Rapidly-exploring Random Tree (RRT)~\cite{RRT_2001} and a particle-based representation to address the uncertainty in state transition. In contrast, we aim to control cables for tasks in which we may be unable to rely on assumptions in~\cite{yamakawa2012simple,kim2016using} for cable motion.

\begin{figure*}[t]
\center
\includegraphics[width=1.00\textwidth]{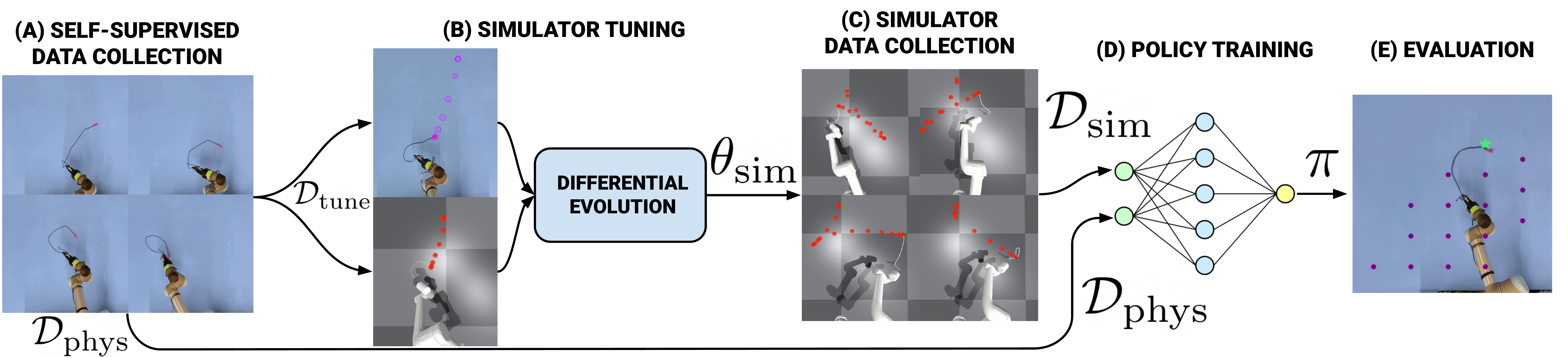}
\caption{The \emph{Real2Sim2Real} pipeline for PRC. We collect a physical dataset $\dreal$ (A) in a self-supervised manner. We subsample $\dreal$ to generate $\dtune$, and use it to tune simulation parameters so that its trajectories match real trajectories using Differential Evolution (B), then use the tuned simulator to generate a large dataset $\dsim$ (C). We use a weighted combination of $\dsim$ and $\dreal$ to train the policy (D) and evaluate the policy in real (E).
}
\vspace*{-10pt}
\label{fig:process}
\end{figure*}

In closely related prior work,~\citet{harry_rope_2021} propose a self-supervised learning technique for dynamic manipulation of fixed-end cables.
In contrast, we use free-end cables. \citet{zimmermanndynamic} study the dynamic manipulation of deformable beams and cloth in the free-end setting. They model elastic objects using the finite-element method~\cite{bathe2006finite} and use optimal control techniques for trajectory optimization. 
The simulated motions perform well in real for simple dynamical systems 
but performance deteriorates for complex soft bodies due to the reality gap. 
We develop a learned, data-driven approach for robotic manipulation of free-end cables and focus on PRC.
\citet{9561766} develop six customizable benchmark simulation environments for 1D deformable objects, but use an commercial closed-source physics engine. We use easily-accessible physics engines to create dynamic simulation environments.

\section{Problem Statement}\label{sec:PS}

In Planar Robot Casting, a robot gripper holds a cable at one endpoint and swings it along a planar surface with a single continuous motion so that the other endpoint comes to rest at a target $\state_d$.
The held cable endpoint is at polar coordinate ${\state=(r, \theta)}$, with the robot base at the origin.
The objective is to find a per-cable policy $\pi$ that minimizes the expected error $||\state_{f, c}-\state_{d, c}||_2$, where $\state_{f, c}$ is the final state and $\state_{d, c}$ is $\state_d$ in Cartesian coordinates. We assume that the $\state_d$ is reachable by the cable endpoint (see Fig.~\ref{fig:teaser}).

\section{Method} \label{sec:approach}

To learn a policy that accounts for variations in cable properties such as mass, stiffness, and friction, 
we propose the \emph{Real2Sim2Real} (R2S2R) framework (Fig.~\ref{fig:process}) and apply it to PRC. To support this framework, we define a reset procedure to bring the system into a consistent starting state and a parameterized trajectory function that generates a dynamic arm action. R2S2R first autonomously collects physical trajectories, then tunes a simulator to match the physical environment and generate simulated trajectories. 
Finally, R2S2R uses a weighted combination of the simulated and physical datasets to train a policy.

\subsection{Reset Procedure}\label{ssec:resets}

To automate data collection and bring the cable into a consistent starting state before each action, we define a 5-step reset procedure, in which the robot
%
(1) lifts the cable up with the free-end touching the surface to prevent the cable from dangling;
(2) continues to lift the cable such that the free-end is just above the surface;
(3) hangs still for 3 seconds to stabilize; 
(4) swings the cable out of the plane, so that the cable straightens along the center axis of the surface and lands with its endpoint far from the robot base; and
(5) slowly pulls the cable towards the robot to the reset position $(r_0, 0)$. 
The reset motion is designed to minimize empirical uncertainty in the reset state, so we assume the start state to be consistent across trajectories. The project website illustrates the reset procedure.

\begin{figure}[t]
\center
\includegraphics[width=0.33\textwidth]{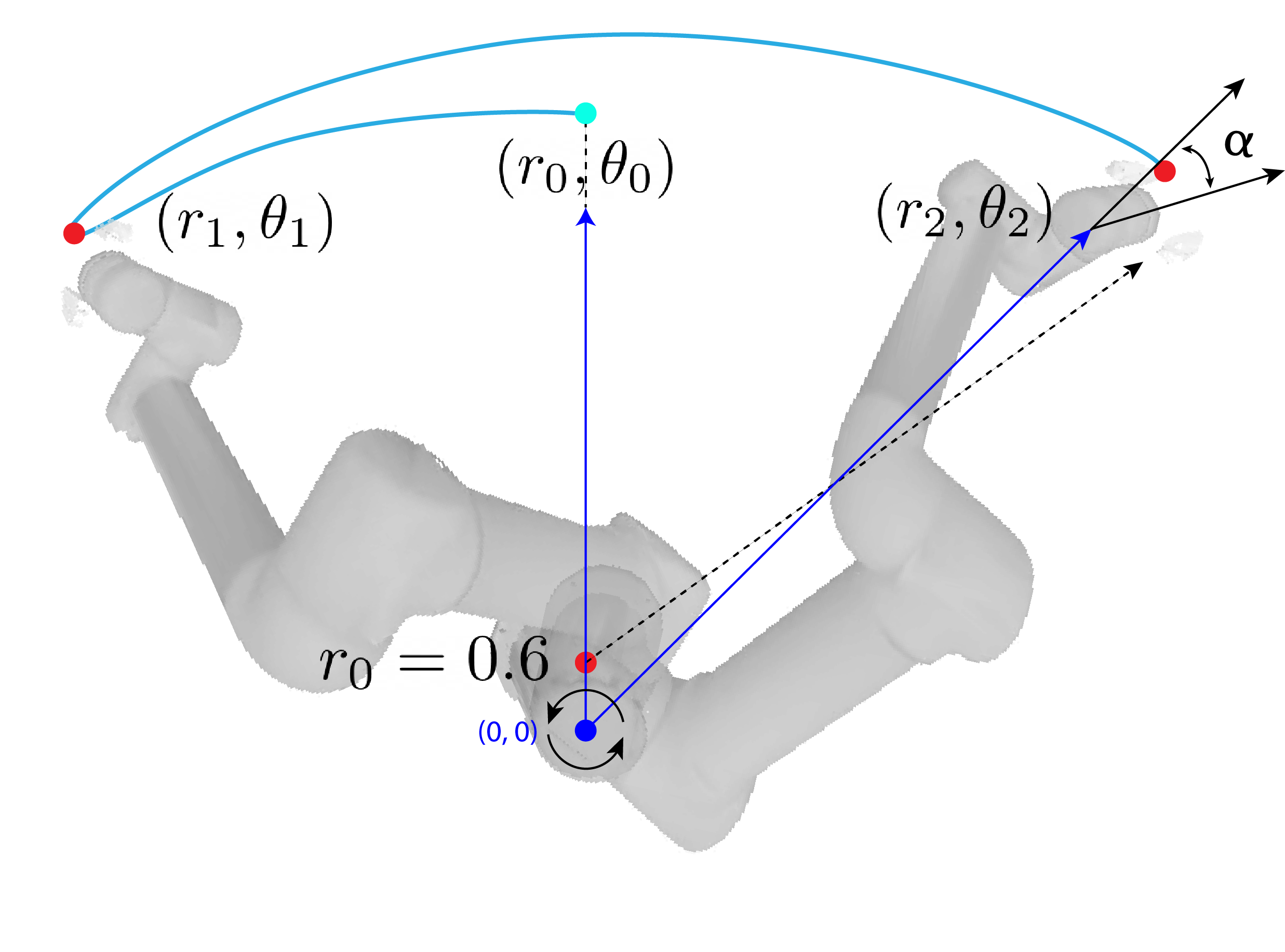}
\vspace{-0.5cm}
\caption{
Example of a spline trajectory traced by the end effector of the UR5 for $r_0=0.6$. The reset procedure brings the end effector to $(r_0, 0)$ and the motion is smoothly interpolated along one spline to $(r_1, \theta_1)$, then along a second spline to $(r_2, \theta_2)$. Along the second spline, an offset $\alpha$ is added to the wrist joint angle produced by the IK solver.
}
\vspace*{-14pt}
\label{fig:spline}
\end{figure}

\subsection{Parameterized Actions}\label{ssec:action}

To generate quick, smooth, dynamic actions that are low dimensional, thus facilitating data generation and training, we define a parameterized action $\ba := (\theta_1, r_1, \theta_2, r_2, \alpha, \vmax)$ composed of two sweeping arcs (Fig.~\ref{fig:spline}).
The motion starts at $(r_0, 0)$, arcs to $(r_1, \theta_1)$, and arcs back to $(r_2, \theta_2)$.
In the motion, $\alpha$ is the wrist joint rotation about the $z$-axis during the second arc, and $\vmax$ is the maximum velocity.
This parameterization is motivated by observing human attempts at the PRC task. 

We convert action parameters to a trajectory in polar coordinates using a cubic spline to smoothly interpolate the radial coefficient from $r_0$ to $r_2$, and use a maximum-velocity spline to interpolate the angular coefficient from $0$ to $\theta_1$ and from $\theta_1$ to $\theta_2$, with maximum velocity $\vmax$. 
The UR5 has difficulty following high-jerk trajectories, so the maximum-velocity spline uses jerk-limited bang-bang control~\cite{ichnowski2020djgomp}. We assume a direction change between the two arc motions, so the angular velocity at $\theta_1$ is 0. 
We use an analytic inverse kinematics (IK) solver to convert the splines to joint configurations. The execution for each trajectory is about 22 seconds, including  20 seconds for the reset motion and 2 seconds for the planar motion.

This parameterization enables state symmetry. For all datasets, we sample actions such that $\theta_1 > 0$, $\theta_2 < 0$, and $\alpha \geq 0$, to obtain targets on the left of the workspace axis of symmetry. If $(\theta_1, \theta_2, r_2, \alpha, \vmax)$ produces target $(r_d, \theta_d)$, $(-\theta_1, -\theta_2, r_2, -\alpha, \vmax)$ will produce $(r_d, -\theta_d)$. Thus, we do not evaluate targets on the right of the axis of symmetry.



\subsection{Self-Supervised Physical Data Collection} \label{ssec:phys-data}

For the first step of R2S2R, we autonomously collect a physical dataset $\dreal$ and take a subset of $\dreal$ to form a simulator tuning dataset $\dtune$ to perform system identification in simulation (Sec.~\ref{ssec:sim-tuning}). 
To create $\dreal$, we grid sample each action parameter to generate $5{\times}5{\times}5{\times}4{\times}2$ trajectories, then filter out trajectories in collision or exceeding joint limits,
to generate $|\dreal|=522$ trajectories. Refer to project website for grid sampling details. 

For automatic data labeling, we record each trajectory and use OpenCV contour detection  \cite{opencv_library} to track a brightly colored endpoint. We extract the 2D waypoint location $p_t = (x_t, y_t)$, where $t$ is the timestep, every 100\,ms from the start of the robot's trajectory. For each trajectory $m_j$ in $\dreal$, the number of waypoints collected is $K_j = \lfloor {T_j}/{100} \rfloor$, where $T_j$ is the duration of $m_j$ in milliseconds. 

\subsection{Three Simulation Models for Robot Casting}\label{ssec:simulators}

\begin{figure}[t]
  \center
  \includegraphics[width=0.46\textwidth]{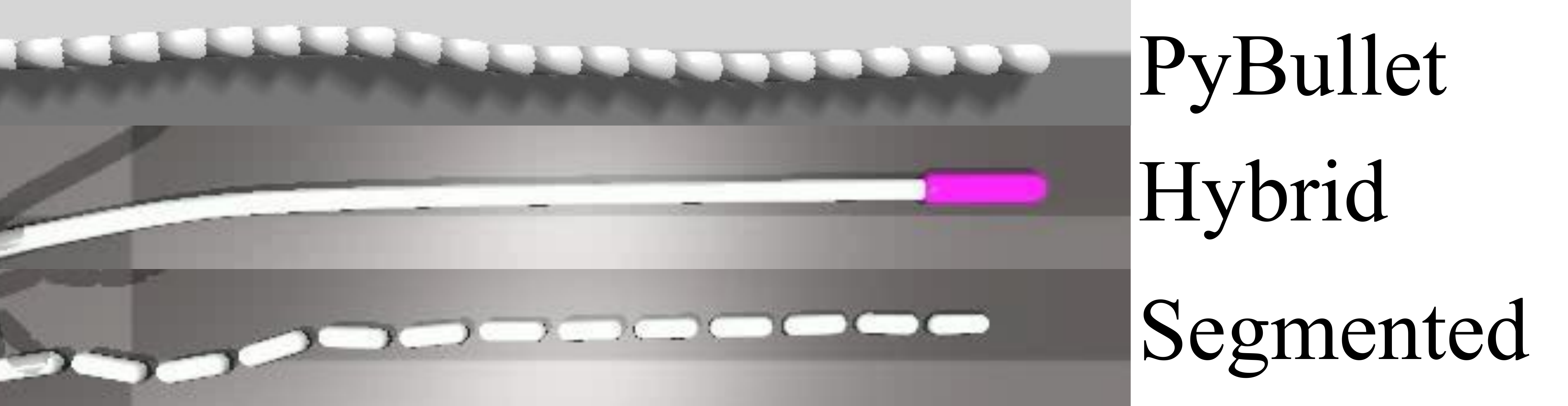}
  \caption{
  Three cable simulation models for Real2Sim tuning. From top to bottom: the PyBullet model consists of capsule rigid bodies connected by 6 DOF spring constraints. The hybrid model is a soft-body rod connected to a capsule rigid body at the endpoint. The segmented model also consists of a string of capsule rigid bodies, but is connected using ball joints.
  }
  \vspace*{-14pt}
  \label{fig:sim_cables}
\end{figure}

R2S2R then tunes a simulator to generate data to augment policy training. We consider which simulator and model best match real from 3 options: PyBullet~\cite{coumans2019} and two versions of NVIDIA Isaac Gym~\cite{makoviychuk2021isaac}.
We set simulated cable geometry (\eg length and radius) to match the real cable. Fig.~\ref{fig:sim_cables} shows a visual comparison, and the website provides further details.

\textbf{PyBullet} is a CPU-based deterministic rigid-body physics simulator used in prior work on deformable object manipulation~\cite{sim2real_deform_2018,seita_bags_2021}. We model the cable as a string of capsule-shaped rigid bodies with 6-DOF spring constraints between each consecutive pair. 
We tune ten parameters: twist stiffness, bend stiffness, mass, lateral friction, spinning friction, rolling friction, endpoint mass, linear damping, angular damping, and dynamic friction. 

\textbf{NVIDIA Isaac Gym} is a GPU-based robotics simulation platform that supports the FleX~\cite{flex_2014} particle-based physics simulator designed for 
deformable objects and rigid bodies. We test two simulation models for the cable: a \emph{segmented} model and a \emph{hybrid} model, both using Flex. 
The segmented model is a string of 18 capsule-shaped rigid bodies with consecutive pairs linked together by 
a ball joint. To model cable stiffness, we tune the joint friction, cable mass, endpoint mass, and planar friction to capture variation in each respective parameter in real.
The hybrid model has the same rigid endpoint from the segmented model, but the rest of the cable is a soft-body rod. 
We tune the Young's modulus of the soft-body rod to model both the cable stiffness and elasticity, dynamic friction of the ground plane, and rigid endpoint mass to capture variations in friction and endpoint properties. Simulation environment details are on the website.

\subsection{Simulator Parameter Tuning for $\dsim$ collection}\label{ssec:sim-tuning}

To tune the parameters of each simulator so that simulated and real trajectories closely match, we follow system identification~\cite{sys_id} and employ a consistent protocol that uses physical trajectories from $\dtune$ (Sec.~\ref{ssec:phys-data}) to minimize the discrepancy between simulated and real trajectories. We define the simulation tuning objective as finding simulation parameters that minimize the average $L^2$ distance between the cable endpoint in simulation $\hat{p_t}$ to the target endpoint $p_t$ over all trajectory timesteps, averaged over a batch of trajectories. We call this metric the average waypoint error. 

As the simulation models may have parameters that have no physical analog, such as the joint friction in the segmented model, or have parameters that do not act similarly to their physical counterparts~\cite{9340852}, we choose tuning algorithms that make no assumptions about the underlying dynamics model. We evaluate two derivative-free black box optimization algorithms, Bayesian Optimization (BO) and Differential Evolution (DE). 

BO builds a probabilistic surrogate model and uses an acquisition function that leverages the uncertainty in the posterior to decide where to query next~\cite{frazier2018tutorial}. BO has been used in policy search for RL hyperparameters~\cite{shahriari2015taking,lizotte2007automatic}, and for tuning simulation fluid dynamics~\cite{9340852}.
DE is an evolutionary population-based stochastic optimization algorithm widely used for global optimization of nondifferentiable, multi-modal, and nonlinear objectives~\cite{storn1997differential}. DE maintains a population of candidate solutions at each iteration and generates new candidates by combing each population member with another mutated population candidate member. It evaluates the fitness of the trial candidates to update the population of candidate solutions until convergence. DE has been shown to be effective at simulation parameter tuning~\cite{de_tuning_2020}.

We evaluate each algorithm via Sim2Sim tests. Using a fixed set of simulation parameters, we generate a small dataset of trajectories to tune randomly initialized simulators to test the effectiveness of DE and BO. We consider the discrepancy between ground truth and predicted simulation parameters and average waypoint error (see results in Sec.~\ref{ssec:sim2sim-results}). We then tune the simulator parameters for each simulation model to the physical data using the best performing tuning algorithm. We tune the simulator using the same process but with \emph{physical} trajectories from $\dtune$.

\subsection{Size of Simulator Tuning Set $\dtune$}
We subsample 20 trajectories from $\dreal$ to generate dataset $\dtune$. We evaluated whether increasing the size of $\dtune$ would reduce the discrepancy between simulation and real using a test set of 30 random physical trajectories not in $\dreal$. We observed negligible difference in test errors between using 20 and 60 trajectories, but the latter require $2\times$ as long to tune on the hybrid and PyBullet models. Tuning with $|\dtune|=20$ trajectories did not lead to a significant speedup in the segmented model, therefore we held this value fixed. 

After tuning, R2S2R generates training data $\dsim$ by grid sampling $\theta_1$, $\theta_2$, $r_2$, $\psi$, and $\vmax$ to generate $15\times15\times15\times10\times2$ values for each respective parameter. We then filter out any trajectories that violate joint limits or with collisions to generate $|\dsim| = 21,450$ $(\ba, \bs)$ simulated trajectories.

\subsection{Policies}\label{ssec:method}

\subsubsection{Forward Dynamics Model}
\label{sssec:supervised}

Due to the multi-modality between a target endpoint $\bs$ and valid actions $\ba$, we do not learn a policy that directly predicts $\ba$ given $\bs_d$. Instead, we learn a forward dynamics model $f_{\rm forw}$ that predicts $\bs$ given $\ba$, and interpolates to select an action. We parameterize $f_{\rm forw}$ using a fully connected neural network. 
Given a dataset $\mathcal{D}$ of $(\ba, \bs_f)$ trajectories, the neural network learns to predict $\bs_f$ given $\ba$ via standard supervised regression. During evaluation, the policy $\pi$ grid samples 67,500 input actions $\ba$ to form a set $\mathcal{A}$ of candidate actions. It then passes each action through the forward dynamics model $f_{\rm forw}$, which outputs the predicted endpoint location $(\hat{x}, \hat{y})$ in Cartesian space. Given a target endpoint $\bs_d$ (in polar coordinates), the policy selects the action that minimizes the Euclidean distance between the predicted endpoint $\bs_f$ and target endpoint $\bs_d$: $\pi(\bs) = \argmin_{\ba \in \mathcal{A}} \; \| \bs_{f, c} - \bs_{d, c} \|_2$, 
where $\state_{f, c}$ and $\state_{d, c}$ are $\state_f$ and $\state_d$ in Cartesian coordinates respectively. Training details can be found on the website. 

\subsubsection{Baseline Policies}\label{sssec:baseline}

We consider two alternative policies:

\textbf{\polarcasting}. Given a target cable endpoint location $\bs_d = (r_d,\theta_d)$, the robot rotates $\theta_d$ radians and executes a rotated reset motion as in Sec.~\ref{ssec:resets}, causing the free endpoint to land at $(r, \theta_d)$. The robot then slowly pulls the cable toward the base for a distance of $r - r_d$. The robot arm is limited to a minimum $r$ coordinate of $r_{\rm min}=0.55$ to prevent the end effector from hitting the robot's supporting table, where $r_{\rm min}$ is the distance from the center of the robot base to the edge of the table, limiting the reachable workspace. 

\textbf{Forward Dynamics Model using Gaussian Process.} As a learning baseline, we train a GP regressor~\cite{Gaussian_process_ML,Mukadam_2018} using $\dreal$ to predict the cable endpoint location $\bs_f$ given input action $\ba$. As with the neural network forward dynamics model, we grid sample 67,500 input actions and select the trajectory that minimizes the predicted Euclidian distance to the target. We do not train a GP regressor using the simulated dataset as GP regression becomes intractable for large datasets with thousands of datapoints~\cite{belyaev2014exact}.

\section{Experiments}
\label{sec:experiments}

\begin{figure}[t]
  \center
  \includegraphics[width=0.45\textwidth]{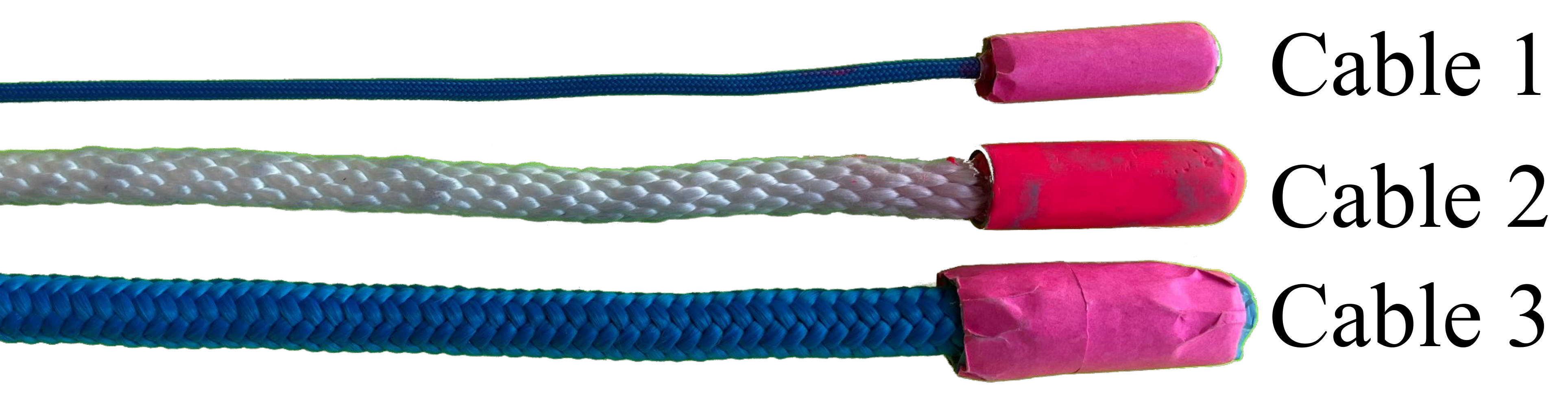}
  \caption{
  Three cables used in physical experiments. Cable 1 is a thin blue paracord, Cable 2 is a nylon cable, and Cable 3 is a thick jump rope. Each endpoint has an attached mass. The respective cable lengths are \SI{0.63}{\meter}, \SI{0.65}{\meter}, and \SI{0.65}{\meter}, the respective masses are \SI{8}{\gram}, \SI{50}{\gram}, and \SI{45}{\gram}, and the respective radii are \SI{4.5}{\milli\meter}, \SI{10}{\milli\meter}, and \SI{14}{\milli\meter}.
  }
  \vspace*{-10pt}
  \label{fig:cables}
\end{figure}

We evaluate the R2S2R pipeline on PRC with 3 cables (Fig.~\ref{fig:cables}) using a UR5 robot (Fig.~\ref{fig:teaser}).
The working surface is a \SI{2.45}{\meter} wide and \SI{1.55}{\meter} tall masonite board, painted blue and sanded to create a consistent friction coefficient across the surface. We use an overhead Logitech Brio 4K webcam recording 1920$\times$1080 images at 60 frames per second.


\begin{table}[t]
  \setlength\tabcolsep{5.0pt}
  \centering
  \footnotesize
  \begin{tabular}{@{}lrrrrrr@{}}
  \toprule
  & \multicolumn{2}{c}{Cable 1}  & \multicolumn{2}{c}{Cable 2} & \multicolumn{2}{c}{Cable 3}\\
  \cmidrule(lr){2-3} \cmidrule(lr){4-5} \cmidrule(l){6-7}
  Sim. Model  & Wayp. & Last & Wayp. & Last & Wayp. & Last \\
  \midrule
  PyBullet         & 29\%  & 28\% & 28\%  & 28\%  & 23\% & 17\% \\
  Isaac Gym Hybrid & 14\%  & 23\% & 11\%  & 14\%  & 11\% & 13\%\\
  Isaac Gym Segm.  & \textbf{9\%} & \textbf{13\%} & \textbf{8\%}  & \textbf{9\%}  & \textbf{11\%} & \textbf{13\%}\\
  \toprule
  \end{tabular}
  \caption{
  Error for simulators tuned using DE on a set of 30 test trajectories for 3 cables (see Fig.~\ref{fig:cables}). The two metrics are the median final $L^2$-distance, which is the 2D Euclidean distance between endpoint locations in simulation and reality after a trajectory terminates, and average waypoint (``Wayp.'') error. Values are expressed in percentages of cable length.
  }
  \vspace*{-10pt}
  \label{tab:tuning}
\end{table}  

We consider three forward dynamics model policies:
\begin{enumerate}
    \item $\pireal$, trained on a small real dataset $\dreal$,
    \item $\pisim$, trained on a large simulated dataset $\dsim$,
    \item $\pitune$, trained on the combined dataset $\dreal \cup \dsim$.
\end{enumerate}
Since $|\dsim| \gg |\dreal|$, we 1) upsample $\dreal$ by randomly duplicating samples such that the real data make between 30\% and 40\% of the combined dataset to force the model to learn more from the real data, and 2) weight the loss function so that samples from $\dreal$ are weighted higher. We empirically choose the upsampling percentage and real sample weights to be the best performing values. 

\begin{figure*}[!h]
    \centering
    \includegraphics[width=\textwidth]{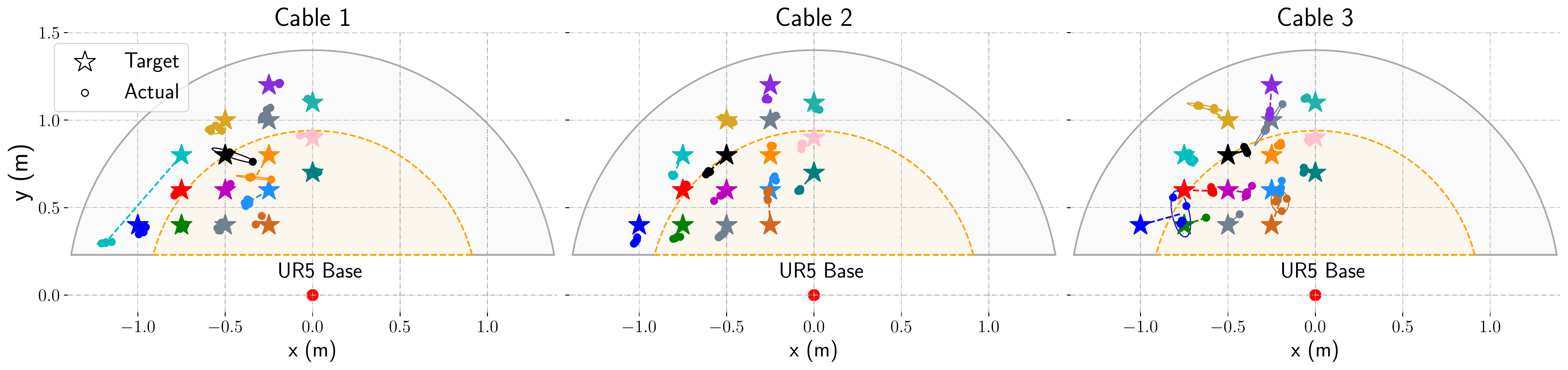}
    \caption{
    Evaluation results for $\pitune$ for each cable. The inner orange shaded region represents the robot workspace and the outer grey shaded region represents the cable workspace.
    The colored stars and dots represent each of 16 target locations and resulting endpoints for 5 trials per target respectively, with associated 95\% confidence ellipses. The dashed lines connect the center of each ellipse to the target location representing the mean error in position. 
}
\label{fig:repeat_multi_cable}
\vspace*{-10pt}
\end{figure*}

\subsection{Comparing Real2Sim Tuning Methods}\label{ssec:sim2sim-results}
To test the effectiveness of BO and DE for tuning the Isaac Gym simulators, we generate simulated data for several hybrid models and segmented models with arbitrarily chosen parameters. We then apply BO and DE to tune the respective simulator model parameters.
For BO, we use the GPyOpt~\cite{gpyopt2016} library, and test three acquisition functions: Expected Improvement (EI), Lower Confidence Bound (LCB), and Maximum Probability of Improvement (MPI).
We use DE as implemented in SciPy~\cite{2020SciPy-NMeth}.
With 5 tuning trajectories, EI performs better than LCB and MPI. The tuning errors using BO range from 3.39\% to 17.98\% for the hybrid model and 1.09\% to 9.25\% for the segmented model. DE consistently tunes the parameters to within $1\%$ of the ground truth parameters for both the hybrid and segmented models, thus we use DE for all further simulator tuning results.

\subsection{Real2Sim}\label{ssec:real2sim-results}
Table~\ref{tab:tuning} reports the Real2Sim simulator tuning results using DE. We observe that the segmented model consistently outperforms the PyBullet and hybrid models in minimizing the discrepancy between simulation and real. For the hybrid model, We find that the tuning algorithms reduce the value of Young's modulus to decrease cable stiffness, but also increases the stretchiness along the length of the cable, which we did not observe in real. We attribute the PyBullet model performance to spring-system simulation instability associated with high spring forces, particularly when preventing the cable from stretching along the length of the cable.

For cable 3, we observe that the hybrid and segmented models perform similarly. Since cable 3 is stiffer, DE produces higher Young's modulus values that both make the simulated cable stiffer and cause it to stretch less along its length, both of which are desired properties.

While the errors for cable 2 are below 10\%, we observe that cables 1 and 3 are harder for each simulator to model, each with a 4\% higher last endpoint error for the segmented model. This may be due to plastic deformation in cables 1 and 3, as we observed that slight bends in the cables would remain until manually straightened by a human operator. We generate $\dsim$ using the segmented model, which requires approximately 4.5 minutes using 4 NVIDIA V100s.

\begin{table}[t]
  \setlength\tabcolsep{5.0pt}
  \centering
  \vspace{1pt}
  \centering
  \scriptsize
  \begin{tabular}{@{}lcrrrrr@{}}
  \toprule
  Model & Dataset & Median & $Q_1$ & $Q_3$ & Min & Max \\
  \midrule
  \polarcasting & N/A & 61\% & 38\% & 86\% & 6\% & 124\%\\
  Gaussian Proc. & $\dreal$ & 27\%  & 9\%   & 51\% & 4\% & 97\%\\
  $\pireal$ & $\dreal$ & 15\% & 11\% & 21\% & 8\% & \textbf{36\%}\\
  $\pisim$ & $\dsim$ & 14\% & 10\% & 17\% & 6\%& 115\%\\
  $\pitune$ & $\dreal \cup \dsim$ & \textbf{8\%} &\textbf{5\%} & \textbf{12\%} & \textbf{2\%} & 105\% \\
  \toprule
  \end{tabular}

    \caption{
    Physical evaluation results on Cable 1 of the 2 baselines and the dynamics model $f_{\rm forw}$ trained on three different datasets over a total of $16\times5=75$ trials. $Q_1$ and $Q_3$ are the first and third quartile. All errors are expressed as a percentage of cable length (\SI{0.65}{\meter}).
  }
  \label{tab:error}
  \vspace*{-5pt}
\end{table}
\subsection{Physical Experiments}
\label{ssec:physical-evaluation}

\begin{table}[t]
  \setlength\tabcolsep{5.0pt}
  \centering
   \small
     \centering
   \begin{tabular}{@{}lrrrrr@{}}
   \toprule
   Cable      & Median & 1st Quartile & 3rd Quartile  & Min & Max   \\
   \midrule
   Cable 1   & 8\% & 5\% & 12\% & 2\% & 105\% \\
   Cable 2   & 12\% & 8\% & 16\%& 4\% & 28\%\\
   Cable 3   & 14\% & 9\% & 23\% & 6\% & 38\% \\
   \toprule
   \end{tabular}
   \caption{
  Physical evaluation results using $\pitune$ of 3 cables (see Fig.~\ref{fig:cables}) on 16 target locations with 5 trials each per cable for a total of 75 trials using policies trained according to $\pitune$. All errors are expressed as percentages of cable length.
   }
   \vspace{-14pt}
  \label{tab:results}
\end{table}

As summarized in Table \ref{tab:error}, we evaluate two baseline policies and PRC policies $\pireal$, $\pisim$, $\pitune$ on cable 1. 
\subsubsection{Baseline Policies} The analytic ``\polarcasting'' baseline performs the worst out of any of the policies, as the robot would collide with the base for targets near the base. The GP trained on $\dreal$ has substantially worse median, 3rd quartile, and maximum errors, but slightly outperforms $\pireal$ in minimum and 1st quartile errors.
\subsubsection{Forward Dynamics Models} We find that policies $\pireal$ and $\pisim$ have similar median and first and third quartile errors, but $\pisim$ has a substantially higher maximum error, corresponding to trajectories where the simulator has high tuning error.
As the simulator is tuned with only 20 trajectories, compared to the over 500 trajectories used to train $\pireal$, we achieve similar evaluation performance while using 96\% fewer physical trajectories. When we combine the two datasets to train $\pitune$, the median error drops by nearly 50\%. The maximum error rose to above 100\% for a single target, suggesting that while the combined dataset substantially improves overall results, it may introduce outliers in regions of the workspace with high reality gap. 
Given that $\pitune$ performs the best out of the policies, we proceed to evaluate the performance for 3 cables on 16 target positions using policy $\pitune$ trained for each cable separately, and repeat each action 5 times per target. Results are summarized in Table~\ref{tab:results}. We quantify the aleatoric uncertainty with the 95\% confidence ellipses in Fig.~\ref{fig:repeat_multi_cable}. 

We attribute the epistemic uncertainty to two sources: the Sim2Real gap, where trajectories executed in simulation do not accurately reflect real, and the learning error in the forward dynamics model. The results suggest that the R2S2R pipeline can apply to other cables.

\section{Conclusion and Future Work}\label{sec:conclusion}

In this paper, we present Real2Sim2Real, a self-supervised learning framework, and apply it to Planar Robot Casting. Experiments suggest that the framework, which collects physical data, tunes a simulator to match the real data, and trains policies from a weighted combination of real and simulated data, can achieve an median error between 12\% and 15\% of cable length on the PRC task.

As motion is restricted to a 2D plane, we do not address the 3D analog of PRC, Spatial Robot Casting (SRC), used in tasks such as fly fishing~\cite{fly_fishing_2004}. PRC is easier to visualize than SRC but can be harder to model due to inherent uncertainty about static and dynamic friction. In both PRC and SRC, the cable is infinite-dimensional and there is a long time horizon for the motion of the cable free end resulting from a motion at the controlled end. 

In future work, we will explore alternate Real2Sim and Sim2Real techniques~\cite{auto-tune_2021,allevato2020tunenet,Fox-RSS-19,dynamics_randomization_2018} and apply Real2Sim2Real to other tasks such as grasping and fabric manipulation.
\section*{Acknowledgments}

{\footnotesize
This research was performed at the AUTOLAB at UC Berkeley in affiliation with the Berkeley AI Research (BAIR) Lab and the CITRIS ``People and Robots'' (CPAR) Initiative. The authors were supported in part by the Toyota Research Institute. We thank our colleagues Justin Kerr, Alejandro Escontrela, Ashwin Balakrishna, Ellen Novoseller, Brijen Thananjeyan, Harry Zhang, Zachary Tam, Chung Min Kim, and Ananth Rao for helpful comments.
}

\renewcommand*{\bibfont}{\footnotesize}
\printbibliography

@String { icra    = {IEEE International Conference on Robotics and Automation (ICRA)} }

@String { ieeera  = {IEEE Robotics and Automation Letters (RA-L)} }

@String { ijrr    = {International Journal of Robotics Research (IJRR)} }

@String { iros    = {IEEE/RSJ International Conference on Intelligent Robots and Systems (IROS)} }

@String { rss     = {Robotics: Science and Systems (RSS)} }

@String { case    = {IEEE Conference on Automation Science and Engineering (CASE)} }

@String { iclr    = {International Conference on Learning Representations (ICLR)} }

@String { corl    = {Conference on Robot Learning (CoRL)} }

@article{ichnowski2020djgomp,
  title={Deep learning can accelerate grasp-optimized motion planning},
  author={Ichnowski, Jeffrey and Avigal, Yahav and Satish, Vishal and Goldberg, Ken},
  journal={Science Robotics},
  volume={5},
  number={48},
  year={2020},
  publisher={Science Robotics}
}

@inproceedings{zimmermanndynamic,
  title={{Dynamic Manipulation of Deformable Objects with Implicit Integration}},
  author={Zimmermann, Simon and Poranne, Roi and Coros, Stelian},
  booktitle=ieeera,
  year={2021},
}

@inproceedings{kim2016using,
  title={{Using a Compliant, Unactuated Tail to Manipulate Objects}},
  author={Kim, Young-Ho and Shell, Dylan A},
  booktitle=ieeera,
  year={2016},
}

@inproceedings{auto-tune_2021,
  author    = {Yuqing Du and Olivia Watkins and Trevor Darrell and Pieter Abbeel and Deepak Pathak},
  title     = {{Auto-Tuned Sim-to-Real Transfer}},
  booktitle  = icra,
  Year       = {2021}
}

@inproceedings{seita_bags_2021,
  author    = {Daniel Seita and Pete Florence and Jonathan Tompson and Erwin Coumans and Vikas Sindhwani and Ken Goldberg and Andy Zeng},
  title     = {{Learning to Rearrange Deformable Cables, Fabrics, and Bags with Goal-Conditioned Transporter Networks}},
  booktitle  = icra,
  Year       = {2021}
}

@inproceedings{harry_rope_2021,
  author    = {Harry Zhang and Jeff Ichnowski and Daniel Seita and Jonathan Wang and Ken Goldberg},
  title     = {{Robots of the Lost Arc: Learning to Dynamically Manipulate Fixed-Endpoint Ropes and Cables}},
  booktitle  = icra,
  Year       = {2021}
}

@inproceedings{closing_sim2real_2019,
  title={{Closing the Sim-to-Real Loop: Adapting Simulation Randomization with Real World Experience}},
  author={Yevgen Chebotar and Ankur Handa and Viktor Makoviychuk and Miles Macklin and Jan Issac and Nathan Ratliff and Dieter Fox},
  booktitle=icra,
  year={2019},
}

@inproceedings{dynamics_randomization_2018,
  title={{Sim-to-Real Transfer of Robotic Control with Dynamics Randomization}},
  author={Xue Bin Peng and Marcin Andrychowicz and Wojciech Zaremba and Pieter Abbeel},
  booktitle=icra,
  year={2018},
}

@inproceedings{nair_rope_2017,
  author={Ashvin Nair and Dian Chen and Pulkit Agrawal and Phillip Isola and Pieter Abbeel and Jitendra Malik and Sergey Levine},
  title={{Combining Self-Supervised Learning and Imitation for Vision-Based Rope Manipulation}},
  booktitle=icra,
  year={2017},
}

@inproceedings{tying_precisely_2016,
  title={{Tying Knot Precisely}},
  author={Weifu Wang and Devin Balkcom},
  booktitle=icra,
  year={2016},
}

@inproceedings{suturing_autolab_2016,
  title={{Automating Multi-Throw Multilateral Surgical Suturing with a Mechanical Needle Guide and Sequential Convex Optimization}},
  author={Siddarth Sen and Animesh Garg and David V Gealy and Stephen McKinley and Yiming Jen and Ken Goldberg},
  booktitle=icra,
  year={2016}
}

@inproceedings{tight_tolerance_insertion_2015,
  title={{An Online Method for Tight-Tolerance Insertion Tasks for String and Rope}},
  author={Weifu Wang and Dmitry Berenson and Devin Balkcom},
  booktitle=icra,
  year={2015},
}

@inproceedings{yamakawa2012simple,
  title={{Simple Model and Deformation Control of a Flexible Rope Using Constant, High-speed Motion of a Robot Arm}},
  author={Yamakawa, Yuji and Namiki, Akio and Ishikawa, Masatoshi},
  booktitle=icra,
  year={2012},
}

@inproceedings{van_den_berg_2010,
  title={{Superhuman Performance of Surgical Tasks by Robots Using Iterative Learning from Human-Guided Demonstrations}},
  author={Jur van den Berg and Stephen Miller and Daniel Duckworth and Humphrey Hu and Andrew Wan and Xiao-Yu Fu and Ken Goldberg and Pieter Abbeel},
  booktitle=icra,
  year={2010},
}

@inproceedings{knot_planning_2003,
  title={{Knot Planning from Observation}},
  author={T. Morita and J. Takamatsu and K. Ogawara and H. Kimura and K. Ikeuchi},
  booktitle=icra,
  year={2003},
}

@inproceedings{wire_insertion_1997,
  title={{Study of Deformation and Insertion Tasks of Flexible Wire}},
  author={H. Nakagaki and K. Kitagi and T. Ogasawara and H. Tsukune},
  booktitle=icra,
  year={1997},
}

@INPROCEEDINGS{9561766,  author={Laezza, Rita and Gieselmann, Robert and Pokorny, Florian T. and Karayiannidis, Yiannis},  booktitle={2021 IEEE International Conference on Robotics and Automation (ICRA)},   title={ReForm: A Robot Learning Sandbox for Deformable Linear Object Manipulation},   year={2021},  volume={},  number={},  pages={4717-4723},  doi={10.1109/ICRA48506.2021.9561766}}

@inproceedings{swingbot_2020,
  author = {Chen Wang and Shaoxiong Wang and Branden Romero and Filipe Veiga and Edward Adelson},
  title = {{SwingBot: Learning Physical Features from In-hand Tactile Exploration for Dynamic Swing-up Manipulation}},
  year = {2020},
  booktitle = iros,
}

@inproceedings{domain_randomization,
  title={{Domain Randomization for Transferring Deep Neural Networks from Simulation to the Real World}},
  author={Josh Tobin and Rachel Fong and Alex Ray and Jonas Schneider and Wojciech Zaremba and Pieter Abbeel},
  booktitle=iros,
  year={2017},
}

@inproceedings{rope_untangling_2013,
  title = {{Tangled: Learning to Untangle Ropes with RGB-D Perception}},
  author = {Wen Hao Lui and Ashutosh Saxena},
  booktitle = iros,
  year = {2013},
}

@inproceedings{dynamic_knotting_2010,
  title = {{Motion Planning for Dynamic Knotting of a Flexible Rope With a High-Speed Robot Arm}},
  author = {Yuji Yamakawa and Akio Namiki and Masatoshi Ishikawa},
  booktitle = iros,
  year = {2010},
}

@inproceedings{one_hand_knotting_tactile_2007,
  title = {{One-Handed Knotting of a Flexible Rope With a High-Speed Multifingered Hand Having Tactile Sensors}},
  author = {Yuji Yamakawa and Akio Namiki and Masatoshi Ishikawa and Makoto Shimojo},
  booktitle = iros,
  year = {2007},
}

@inproceedings{robot_heart_surgery_2006,
  title = {{A System for Robotic Heart Surgery that Learns to Tie Knots Using Recurrent Neural Networks}},
  author = {Hermann Mayer and Faustino Gomez and Daan Wierstra and Istvan Nagy and Alois Knoll and Jurgen Schmidhuber},
  booktitle = iros,
  year = {2006},
}

@inproceedings{lynch_mason_1993,
  title={{Dynamic Manipulation}},
  author={Matthew T Mason and Kevin M Lynch},
  booktitle= iros,
  year={1993},
}

@INPROCEEDINGS{9340852,  author={Lopez-Guevara, Tatiana and Pucci, Rita and Taylor, Nicholas K. and Gutmann, Michael U. and Ramamoorthy, Suhramanian and Suhr, Kartic},  booktitle={2020 IEEE/RSJ International Conference on Intelligent Robots and Systems (IROS)},   title={Stir to Pour: Efficient Calibration of Liquid Properties for Pouring Actions},   year={2020},  volume={},  number={},  pages={5351-5357},  doi={10.1109/IROS45743.2020.9340852}}

@inproceedings{grannen2020untangling,
  title={{Untangling Dense Knots by Learning Task-Relevant Keypoints}},
  author={Grannen, Jennifer and Sundaresan, Priya and Thananjeyan, Brijen and Ichnowski, Jeffrey and Balakrishna, Ashwin and Hwang, Minho and Viswanath, Vainavi and Laskey, Michael and Gonzalez, Joseph E. and Goldberg, Ken},
  booktitle=corl,
  year={2020}
}

@inproceedings{corl2020softgym,
  title={{SoftGym: Benchmarking Deep Reinforcement Learning for Deformable Object Manipulation}},
  author={Lin, Xingyu and Wang, Yufei and Olkin, Jake and Held, David},
  booktitle=corl,
  year={2020}
}

@inproceedings{yan_fabrics_latent_2020,
  author = {Wilson Yan and Ashwin Vangipuram and Pieter Abbeel and Lerrel Pinto},
  title = {{Learning Predictive Representations for Deformable Objects Using Contrastive Estimation}},
  booktitle = corl,
  Year = {2020}
}

@inproceedings{zeng_transporters_2020,
  title={{Transporter Networks: Rearranging the Visual World for Robotic Manipulation}},
  author={Andy Zeng and Pete Florence and Jonathan Tompson and Stefan Welker and Jonathan Chien and Maria Attarian and Travis Armstrong and Ivan Krasin and Dan Duong and Vikas Sindhwani and Johnny Lee},
  booktitle=corl,
  year={2020}
}

@article{sim2real_deform_2018,
  title={{Sim-to-Real Reinforcement Learning for Deformable Object Manipulation}},
  author={Jan Matas and Stephen James and Andrew J. Davison},
  journal=corl,
  year=2018
}

@inproceedings{allevato2020tunenet,
  title={Tunenet: One-shot residual tuning for system identification and sim-to-real robot task transfer},
  author={Allevato, Adam and Short, Elaine Schaertl and Pryor, Mitch and Thomaz, Andrea},
  booktitle={Conference on Robot Learning},
  pages={445--455},
  year={2020},
  organization={PMLR}
}

@inproceedings{ruiz2011fast,
  title={Fast adaptation for effect-aware pushing},
  author={Ruiz-Ugalde, Federico and Cheng, Gordon and Beetz, Michael},
  booktitle={2011 11th IEEE-RAS International Conference on Humanoid Robots},
  pages={614--621},
  year={2011},
  organization={IEEE}
}

@inproceedings{high_speed_knotting_2013,
  author={Yuji Yamakawa and Akio Namiki and Masatoshi Ishikawa},
  title={{Dynamic High-Speed Knotting of a Rope by a Manipulator}},
  booktitle={International Journal of Advanced Robotic Systems},
  year=2013,
}

@inproceedings{how_to_train_rl,
  author = {Julian Ibarz and Jie Tan and Chelsea Finn and Mrinal Kalakrishnan and Peter Pastor and Sergey Levine},
  title ={{How to Train Your Robot with Deep Reinforcement Learning: Lessons we Have Learned}},
  booktitle = ijrr,
  year = {2021},
}

@inproceedings{manip_deformable_survey_2018,
  title={{Robotic Manipulation and Sensing of Deformable Objects in Domestic and Industrial Applications: a Survey}},
  author={Jose Sanchez and Juan-Antonio Corrales and Belhassen-Chedli Bouzgarrou and Youcef Mezouar},
  booktitle=ijrr,
  year={2018},
}

@inproceedings{Mukadam_2018,
   title={{Continuous-time Gaussian Process Motion Planning via Probabilistic Inference}},
   author={Mukadam, Mustafa and Dong, Jing and Yan, Xinyan and Dellaert, Frank and Boots, Byron},
   booktitle=ijrr,
   year={2018},
}

@inproceedings{RRT_2001,
  title={{Randomized Kinodynamic Planning}},
  author={Steven M LaValle and James J Kuffner},
  booktitle = ijrr,
  year = 2001
}

@inproceedings{case_study_knots_1991,
  title={{A Case Study of Flexible Object Manipulation}},
  author={J. Hopcroft and J. Kearney and D. Krafft},
  booktitle = ijrr,
  year = 1991
}

@article{lynch1999dynamic,
  title={Dynamic nonprehensile manipulation: Controllability, planning, and experiments},
  author={Lynch, Kevin M and Mason, Matthew T},
  journal={The International Journal of Robotics Research},
  volume={18},
  number={1},
  pages={64--92},
  year={1999},
  publisher={SAGE Publications}
}

@inproceedings{tactile_cable_2020,
  title = {{Cable Manipulation with a Tactile-Reactive Gripper}},
  author = {Yu She and Siyuan Dong and Shaoxiong Wang and Neha Sunil and Alberto Rodriguez and Edward Adelson},
  booktitle = rss,
  Year = {2020}
}

@inproceedings{lerrel_2020,
  title={{Learning to Manipulate Deformable Objects without Demonstrations}},
  author={Yilin Wu and Wilson Yan and Thanard Kurutach and Lerrel Pinto and Pieter Abbeel},
  booktitle=rss,
  year={2020},
}

@inproceedings{zeng_tossing_2019,
  title={{TossingBot: Learning to Throw Arbitrary Objects with Residual Physics}},
  author={Andy Zeng and Shuran Song and Johnny Lee and Alberto Rodriguez and Thomas Funkhouser},
  booktitle=rss,
  year={2019},
}

@inproceedings{cad2rl,
  title={{CAD2RL: Real Single-Image Flight without a Single Real Image}},
  author={Fereshteh Sadeghi and Sergey Levine},
  booktitle=rss,
  year=2017,
}

@inproceedings{mahler2017dexnet,
  title={{Dex-Net 2.0: Deep Learning to Plan Robust Grasps with Synthetic Point Clouds and Analytic Grasp Metrics}},
  author={Jeffrey Mahler and Jacky Liang and Sherdil Niyaz and Michael Laskey and Richard Doan and Xinyu Liu and Juan Aparicio Ojea and Ken Goldberg},
  booktitle=rss,
  year={2017},
}

@INPROCEEDINGS{Fox-RSS-19, 
    AUTHOR    = {Fabio Ramos AND Rafael Possas AND Dieter Fox}, 
    TITLE     = {BayesSim: Adaptive Domain Randomization Via Probabilistic Inference for Robotics Simulators}, 
    BOOKTITLE = {Proceedings of Robotics: Science and Systems}, 
    YEAR      = {2019}, 
    ADDRESS   = {FreiburgimBreisgau, Germany}, 
    MONTH     = {June}, 
    DOI       = {10.15607/RSS.2019.XV.029} 
}

@inproceedings{lizotte2007automatic,
  title={Automatic Gait Optimization with Gaussian Process Regression.},
  author={Lizotte, Daniel J and Wang, Tao and Bowling, Michael H and Schuurmans, Dale and others},
  booktitle={IJCAI},
  volume={7},
  pages={944--949},
  year={2007}
}

@inproceedings{ZSVI_2018,
  author = {Deepak Pathak and Parsa Mahmoudieh and Guanghao Luo and Pulkit Agrawal and Dian Chen and Yide Shentu and Evan Shelhamer and Jitendra Malik and Alexei A. Efros and Trevor Darrell},
  title = {{Zero-Shot Visual Imitation}},
  booktitle = iclr,
  year = 2018
}

@book{sys_id,
  author = {Ljung, Lennart},
  title = {System Identification: Theory for the User},
  year = {1986},
  isbn = {0138816409},
  publisher = {Prentice-Hall, Inc.},
  address = {USA}
}

@book{Gaussian_process_ML,
  author = {Carl Rasmussen and Christopher Williams},
  title = {{Gaussian Processes for Machine Learning}},
  year = {2006},
  publisher = {MIT Press},
}

@book{bathe2006finite,
  title={{Finite Element Procedures}},
  author={Bathe, K.J.},
  url={https://books.google.com/books?id=rWvefGICfO8C},
  year={2006},
  publisher={Prentice Hall}
}

@article{shahriari2015taking,
  title={Taking the human out of the loop: A review of Bayesian optimization},
  author={Shahriari, Bobak and Swersky, Kevin and Wang, Ziyu and Adams, Ryan P and De Freitas, Nando},
  journal={Proceedings of the IEEE},
  volume={104},
  number={1},
  pages={148--175},
  year={2015},
  publisher={IEEE}
}

@article{flex_2014,
  author = {Macklin, Miles and Muller, Matthias and Chentanez, Nuttapong and Kim, Tae-Yong},
  title = {{Unified Particle Physics for Real-Time Applications}},
  year = {2014},
  volume = {33},
  number = {4},
  journal = {ACM Trans. Graph.},
  month = jul,
}

@MISC{coumans2019,
  author =   {Erwin Coumans and Yunfei Bai},
  title =    {{PyBullet, a Python Module for Physics Simulation for Games, Robotics and Machine Learning}},
  howpublished = {\url{http://pybullet.org}},
  year = {2021}
}

@misc{makoviychuk2021isaac,
  title={{Isaac Gym: High Performance GPU-Based Physics Simulation For Robot Learning}}, 
  author={Viktor Makoviychuk and Lukasz Wawrzyniak and Yunrong Guo and Michelle Lu and Kier Storey and Miles Macklin and David Hoeller and Nikita Rudin and Arthur Allshire and Ankur Handa and Gavriel State},
  year={2021},
  eprint={2108.10470},
  archivePrefix={arXiv},
  primaryClass={cs.RO}
}

@ARTICLE{2020SciPy-NMeth,
  author  = {Virtanen, Pauli and Gommers, Ralf and Oliphant, Travis E. and
            Haberland, Matt and Reddy, Tyler and Cournapeau, David and
            Burovski, Evgeni and Peterson, Pearu and Weckesser, Warren and
            Bright, Jonathan and {van der Walt}, St{\'e}fan J. and
            Brett, Matthew and Wilson, Joshua and Millman, K. Jarrod and
            Mayorov, Nikolay and Nelson, Andrew R. J. and Jones, Eric and
            Kern, Robert and Larson, Eric and Carey, C J and
            Polat, {\.I}lhan and Feng, Yu and Moore, Eric W. and
            {VanderPlas}, Jake and Laxalde, Denis and Perktold, Josef and
            Cimrman, Robert and Henriksen, Ian and Quintero, E. A. and
            Harris, Charles R. and Archibald, Anne M. and
            Ribeiro, Ant{\^o}nio H. and Pedregosa, Fabian and
            {van Mulbregt}, Paul and {SciPy 1.0 Contributors}},
  title   = {{{SciPy} 1.0: Fundamental Algorithms for Scientific
            Computing in Python}},
  journal = {Nature Methods},
  year    = {2020},
  volume  = {17},
  pages   = {261--272},
  adsurl  = {https://rdcu.be/b08Wh},
  doi     = {10.1038/s41592-019-0686-2},
}

@Misc{gpyopt2016,
  author =   {The GPyOpt authors},
  title =    {GPyOpt: A Bayesian Optimization framework in Python},
  howpublished = {\url{http://github.com/SheffieldML/GPyOpt}},
  year = {2016}
}

@misc{SpinningUp2018,
    author = {Achiam, Joshua},
    title = {{Spinning Up in Deep Reinforcement Learning}},
    year = {2018}
}

@article{opencv_library,
    author = {Bradski, G.},
    journal = {Dr. Dobb's Journal of Software Tools},
    title = {{The OpenCV Library}},
    year = {2000}
}

@ARTICLE{catching_2014,
  author={Kim, Seungsu and Shukla, Ashwini and Billard, Aude},
  journal={IEEE Transactions on Robotics}, 
  title={{Catching Objects in Flight}},
  year={2014},
  volume={30},
  number={5},
  pages={1049-1065},
}

@article{wang2011analysis,
  title={{Analysis of Fly Fishing Rod Casting Dynamics}},
  author={Wang, Gang and Wereley, Norman},
  journal={Shock and Vibration},
  volume={18},
  number={6},
  pages={839--855},
  year={2011},
  publisher={IOS Press}
}

@inproceedings{sim2real2sim_2020,
  title={{Sim2Real2Sim: Bridging the Gap Between Simulation and Real-World in Flexible Object Manipulation}},
  author={Peng Chang and Taskin Padir},
  booktitle={IEEE International Conference on Robotic Computing (IRC)},
  year={2020},
}

@article{storn1997differential,
  title={Differential evolution--a simple and efficient heuristic for global optimization over continuous spaces},
  author={Storn, Rainer and Price, Kenneth},
  journal={Journal of global optimization},
  volume={11},
  number={4},
  pages={341--359},
  year={1997},
  publisher={Springer}
}

@inproceedings{reality_gap_1995,
  title={{Noise and the Reality Gap: The use of Simulation in Evolutionary Robotics}},
  author={N. Jakobi and P.Husbands and I. Harvey},
  booktitle={European Conference on Advances in Artificial Life},
  year={1995},
}

@article{fly_fishing_2004,
  title={{Numerical Model for the Dynamics of a Coupled Fly Line / Fly Rod System and Experimental Validation}},
  author={Caroline Gatti-Bonoa and N.C. Perkins},
  journal={Journal of Sound and Vibration},
  year={2004},
}

@article{de_tuning_2020,
  title={{Traversing the Reality Gap via Simulator Tuning}},
  author={Jack Collins and Ross Brown and Jurgen Leitner and David Howard},
  journal={arXiv preprint arXiv:2003.01369},
  year={2020},
}

@article{diabolo_2020,
  title={{An Analytical Diabolo Model for Robotic Learning and Control}},
  author={Felix von Drigalski and Devwrat Joshi and Takayuki Murooka and Kazutoshi Tanaka and Masashi Hamaya and Yoshihisa Ijiri},
  journal={arXiv preprint arXiv:2011.09068},
  year={2020},
}

@article{frazier2018tutorial,
  title={A tutorial on Bayesian optimization},
  author={Frazier, Peter I},
  journal={arXiv preprint arXiv:1807.02811},
  year={2018}
}

@misc{rss_workshop,
  title = {{RSS 2021 Workshop on Deformable Object Simulation in Robotics}},
  howpublished = {\url{https://sites.google.com/nvidia.com/do-sim}},
  author = {Yashraj Narang and Yunzhu Li and Miles Macklin and Dale McConachie and Jane Wu},
  note = {Accessed: 2021-09-12}
}

@article{belyaev2014exact,
  title={Exact inference for Gaussian process regression in case of big data with the Cartesian product structure},
  author={Belyaev, Mikhail and Burnaev, Evgeny and Kapushev, Yermek},
  journal={arXiv preprint arXiv:1403.6573},
  year={2014}
}

\end{document}